%% file: main.tex
\documentclass[10pt,twocolumn,letterpaper]{article}

\usepackage{cvpr}      

\usepackage{graphicx}
\usepackage{amsmath,bm}
\usepackage{amssymb}
\usepackage{booktabs}
\usepackage{color}
\usepackage{multirow}
\usepackage{threeparttable}
\usepackage{enumitem}
\usepackage{url}
\usepackage{ntheorem}
\usepackage{wrapfig}

\usepackage[pagebackref,breaklinks,colorlinks]{hyperref}

\usepackage[capitalize]{cleveref}
\crefname{section}{Sec.}{Secs.}
\Crefname{section}{Section}{Sections}
\Crefname{table}{Table}{Tables}
\crefname{table}{Tab.}{Tabs.}

\usepackage[font=small,skip=3pt]{caption}
\def\cvprPaperID{1269} 
\def\confName{CVPR}
\def\confYear{2022}
\graphicspath{{./figures/}}

\newtoggle{independent_supp}        
\togglefalse{independent_supp}

\begin{document}

\title{\Large Structured Local Radiance Fields for Human Avatar Modeling}

\author{
	Zerong Zheng\textsuperscript{1},
	Han Huang\textsuperscript{2},
	Tao Yu\textsuperscript{1},
	Hongwen Zhang\textsuperscript{1},
	Yandong Guo\textsuperscript{2},
	Yebin Liu\textsuperscript{1}
	\\ \\
	\textsuperscript{1}Department of Automation, Tsinghua University
	\quad
	\textsuperscript{2}OPPO Research Institute 
}

\maketitle

\begin{abstract}


It is extremely challenging to create an animatable clothed human avatar from RGB videos, especially for loose clothes due to the difficulties in motion modeling. To address this problem, we introduce a novel representation on the basis of recent neural scene  rendering techniques. The core of our representation is a set of structured local radiance fields, which are anchored to the pre-defined nodes sampled on a statistical human body template. These local radiance fields not only leverage the flexibility of implicit representation in shape and appearance modeling, but also factorize cloth deformations into skeleton motions, node residual translations and the dynamic detail variations inside each individual radiance field. To learn our representation from RGB data and facilitate pose generalization, we propose to learn the node translations and the detail variations in a conditional generative latent space. Overall, our method enables automatic construction of animatable human avatars for various types of clothes without the need for scanning subject-specific templates, and can generate realistic images with dynamic details for novel poses. Experiment show that our method outperforms state-of-the-art methods both qualitatively and quantitatively.

\end{abstract}

\input{1_intro}
\input{2_related_work}

\input{3_overview}

\input{4_method}

\input{5_results}

\input{6_conclusion}

{\small
\bibliographystyle{ieee_fullname}
\bibliography{egbib}
}

\clearpage
\renewcommand\thesection{\Alph{section}}
\renewcommand\thefigure{\Alph{figure}}
\renewcommand\thetable{\Alph{table}}
\setcounter{section}{0}
\setcounter{figure}{0}
\setcounter{table}{0}
\noindent\textbf{\Large Supplemental Document} 
\input{9_supp_context}

\end{document}

%% file: 1_intro.tex
\section{Introduction}
\label{sec:intro}

Animatable human avatar modeling is of great importance in many applications such as content creation and entertainment, and virtual characters have become ubiquitous in our lives with the rise of computer graphics in movies and games. Traditional methods for high-quality human avatar reconstruction are often costly and tedious, due to the difficulties in modeling the complex dynamics of clothes. Besides, they typically presume the availability of a subject-specific template~\cite{habermann2021realtimeDDC} and its accurate registration to the input frames~\cite{timur2021driving_signal,Xiang2021ModelingClothing}, which are difficult to acquire in practice.

With the rapid development in computer vision in the past ten years, researchers have started to explore the possibility of automatic human avatar reconstruction without pre-scanning efforts. Pioneer studies deformed a statistical human body template (e.g., SMPL \cite{loper2015smpl}) to model the clothed human geometry and appearance \cite{alldieck2018videoavatar,alldieck2018videoavatar_detailed,alldieck19octopus}. Neural texture maps and image-to-image networks are later adopted to achieve photo-realistic rendering \cite{Liu2018Neural,liu2020NeuralHumanRendering,Shysheya2019TNR,raj2020anr}. Recently, neural radiance representations, which implicitly encode shape and appearance using neural networks, are also applied in pursuit of higher-fidelity results \cite{peng2021animatable_nerf,noguchi2021narf,neural_actors}. These methods typically define the radiance field in a canonical pose, and warp it to live poses using linear blending skinning (LBS) under the guidance of the SMPL surface.

\begin{figure}
    \centering
    \includegraphics[width=1.0\linewidth]{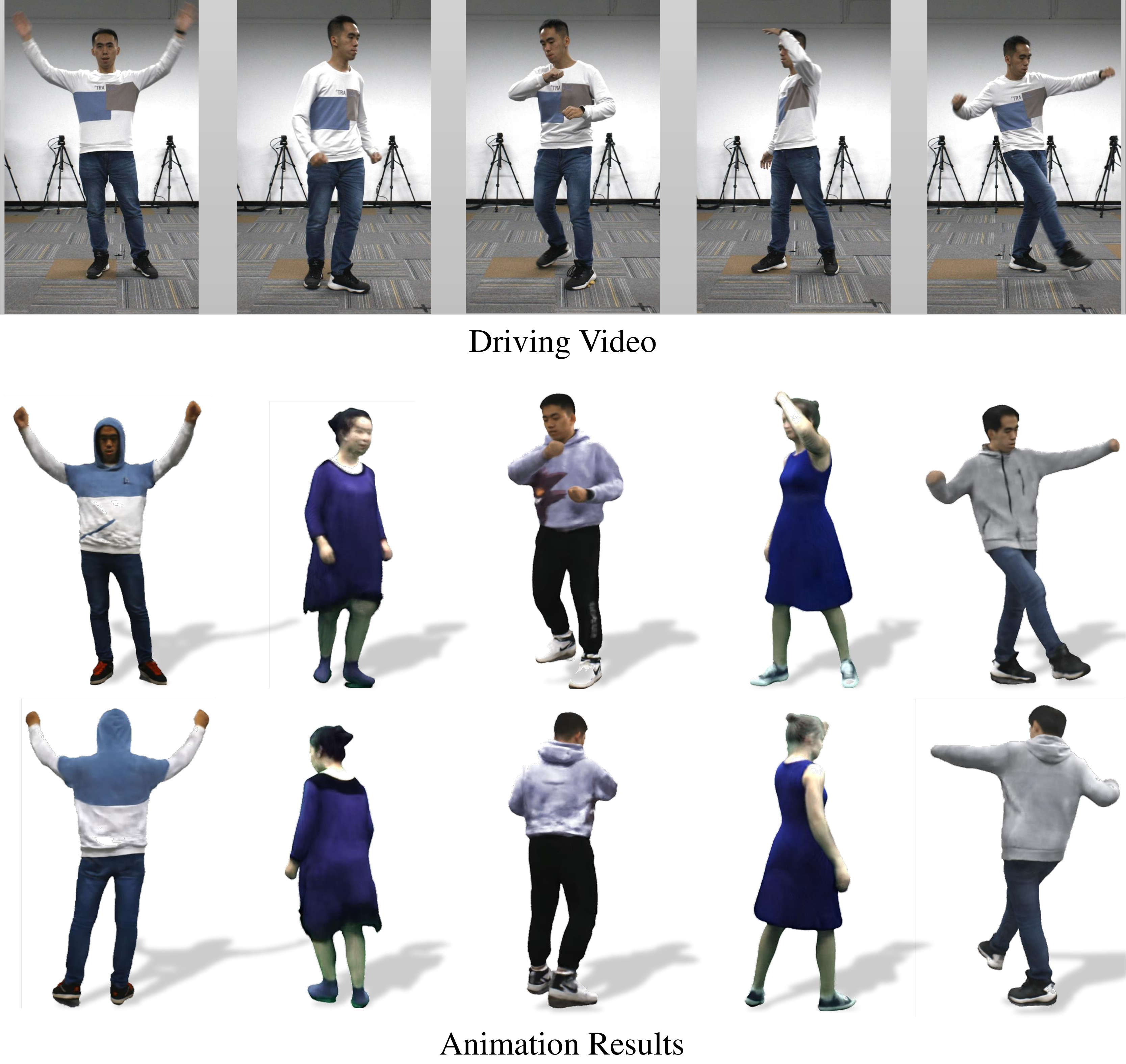}
    \caption{\textbf{Example results produced by our method.} Our method can learn animatable human avatars with various cloth topologies and realistic dynamic details. Top row: driving video, from which the animation poses are extracted. Bottom two rows: animation results rendered from the front and the back view.  }
    \label{fig:teaser}
\end{figure}

Despite the differences in the representations inside the aforementioned approaches, we find that there is one thing in common: they all heavily rely on the skeleton or the surface of SMPL model for cloth motion modeling. This is apparent in methods based on the SMPL topology, either using traditional texture maps \cite{alldieck2018videoavatar,alldieck2018videoavatar_detailed,alldieck19octopus} or neural textures \cite{Liu2018Neural,liu2020NeuralHumanRendering,Shysheya2019TNR,raj2020anr}. Even in state-of-the-art methods based on implicit fields \cite{peng2021animatable_nerf,noguchi2021narf,neural_actors}, researchers still assumed that skin motions can be propagated to approximate the cloth deformations, which, unfortunately, only holds for tight-fitting clothes. 
When applying these methods to loose clothes, articulation motions based on solely body joints cannot express the complete information about the wrinkles and non-rigid deformations. Some methods learned to directly regress cloth deformations from body pose configurations~\cite{neural_actors}; however, the complexity gap between body poses and cloth details results in a one-to-many mapping problem, leading to under-fitting issues where the network learns averaged, blurry appearance. 
Suffering from this fatal limitation, no methods have demonstrated animatable human characters wearing skirts or dresses so far. 

To overcome this limitation and fill the void, we propose a new representation for clothed human characters. Our representation is built upon neural radiance fields~\cite{mildenhall2020nerf}, or NeRF in short, for its excellent performance in learning the appearance of static scenes. To extend NeRF for dynamic character modeling, we break a global NeRF into a set of \emph{structured local radiance fields}, which are attached to the pre-defined nodes on the SMPL model. Each local radiance field is responsible for representing the shape and appearance in the local space around its corresponding node. The local radiance fields can be driven by the body skeleton, while having their own residual movements to represent the non-rigid deformation of garments. Furthermore, each radiance field is conditioned on a dynamic detail embedding, which encodes the high-frequency dynamic details that cannot be modeled via node translation. In this way, our representation decomposes the cloth deformations in a coarse-to-fine manner: the coarsest level is the skeleton motion, the middle level is the residual movements of the local radiance fields, and the finest level is the time-varying details inside each radiance field. 

However, employing such a representation for avatar modeling is not straight-forward as the node-related variables (\textit{i.e.}, the node residual translations and the dynamic detail embeddings) are difficult to acquire in practice. 
Although we can obtain these variables for training frames through naive optimization with image evidence, it remains unclear how to compute them for unseen poses. 
Alternatively, one can train a network that directly regresses these variables from body poses, but this will result into the aforementioned under-fitting issues due to information deficiency~\cite{timur2021driving_signal}. 
In order to achieve a balance between data fitting and generalization, we draw inspiration from \cite{timur2021driving_signal} and learn the node-related variables in a conditional generative latent space. Specifically, we introduce a tiny conditional variational auto-encoder (cVAE)~\cite{Sohn2015cvae} for each local radiance field. 
Conditioned on the pose parameters, the cVAE decoders convert the latent bottlenecks into node-related variables. For the input of the cVAE encoder, we find that the time stamp~\cite{Pumarola20arxiv_D_NeRF,Xian20arxiv_stnif,Gao-freeviewvideo} is an effective option, because it is simple, distinguishable, and naturally guarantees the temporal smoothness of the node-related variables thanks to the low-frequency bias in MLPs~\cite{tancik2020fourfeat}. 
Intuitively, the time stamp is provided as an auxiliary input to help our network distinguish similar poses at different frames, while the VAE property can push the latent space to be uninformative, thereby encouraging the network to mainly rely on pose conditions when inferring node-related variables. 
With all of these building blocks, our network can be trained in an end-to-end manner, eventually producing a realistic dynamic human avatar.








Overall, our proposed method offers the new ability to automatically create an animatable human character with general, dynamic garments. This is achieved by using only RGB videos, without any pre-scanning efforts. Compared to methods that heavily depend on the topology of a naked human body template, our approach is powerful yet general in terms of both appearance learning and motion modeling, and able to generate realistic dynamic details. To the best of our knowledge, our method is the first one that demonstrates automatic human avatar creation for dresses. Experiments prove that our method outperforms state-of-the-art approaches qualitatively and quantitatively. 

%% file: 2_related_work.tex
\begin{figure*}
    \centering
    \includegraphics[width=1.0\linewidth]{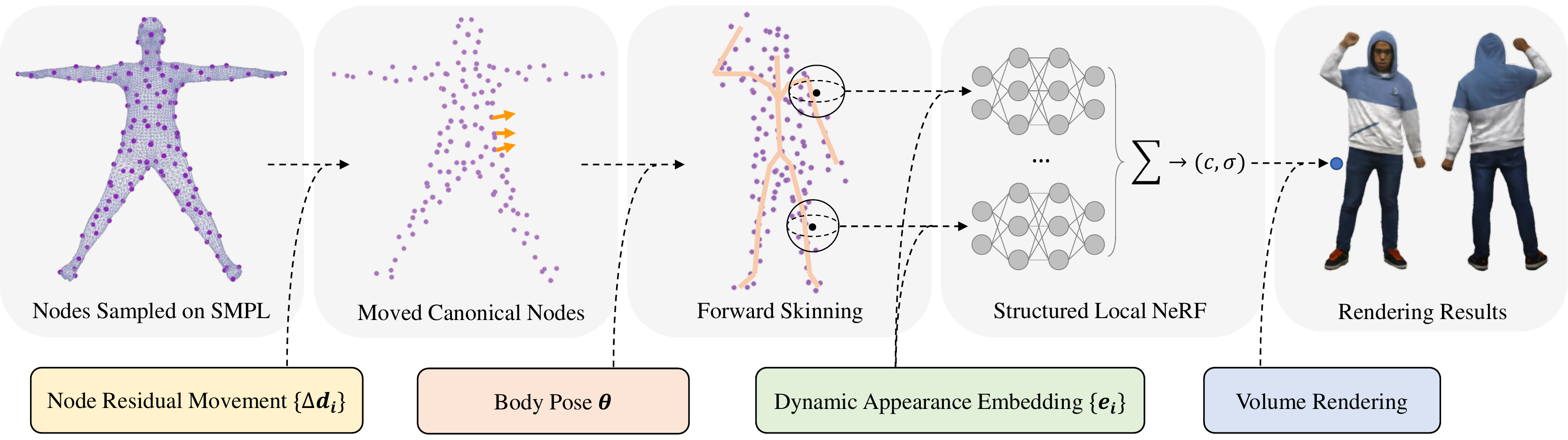}
    \caption{\textbf{Illustration of our clothed human representation.} In our proposed method, we represent the dynamic appearance of a clothed human character using structured local radiance fields attached to pre-defined nodes on the SMPL model. The garment deformations are then modeled in a coarse-to-fine manner with three set of variables, including the body poses as the coarsest level, the node residual translations as the middle level and the dynamic detail embeddings of the local radiance fields as the finest level. }
    \label{fig:reprez}
\end{figure*}

\section{Related Work}
\label{sec:related}

\textbf{Image-based 3D Human Reconstruction.} 
Three-dimensional human character reconstruction is traditionally the very first step towards human avatar modeling. Previous studies focused on using multi-view images~\cite{StarckCGA07,liu2009point,WuShadingHuman,WaschbuschWCSG05,VlasicPBDPRM09} or RGB(D) image sequences~\cite{alldieck2018videoavatar,alldieck2018videoavatar_detailed,yu2017BodyFusion,yu2018doublefusion,Zheng2018HybridFusion,bogo2015detailed,dou2016fusion4d,Motion2Fusion,yu2021function4d,Xiang2020MonoClothCap,habermann2019livecap,Xu2020UnstructuredFusion,guo2017real} for human model reconstruction. Extremely high-quality reconstruction results have also been demonstrated with tens or even hundreds of cameras~\cite{collet2015high}. In order to reduce the difficulty in system setup, human model reconstruction from sparse camera views has been investigated by using neural networks for learning silhouette cues~\cite{MinimalCam18,natsume2019siclope} and stereo cues~\cite{SparseViewHaoLi18}. More recently, various approaches were proposed to reconstruct a 3D human model from a single-view RGB images~\cite{saito2019pifu,saito2020pifuhd,zheng2020pamir,gabeur2019moulding,wang2020normalgan,zhu2019hmd,alldieck2019tex2shape,ARCH2020,He2021archplusplus}. For example, PIFu~\cite{saito2019pifu} and PIFuHD~\cite{saito2020pifuhd} proposed to regress a deep implicit function using pixel-aligned image features and is able to reconstruct high-resolution results. ARCH~\cite{ARCH2020} and ARCH++~\cite{He2021archplusplus} proposed to reconstruct the 3D human model in a canonical pose in order to support animation. Although demonstrating plausible results, these methods rely on large scale dataset of 3D human scans to train the model, and suffer from reconstruction errors and weak generalization capability. In contrast, our method bypasses the reconstruction step and directly learns an animatable avatar from RGB videos.

\textbf{Neural Scene Representations and Rendering.} 
Representing objects or scenes implicitly with neural networks, is becoming more and more popular for its compactness and strong representation power. Pioneer studies proposed to learn an implicit function where the shapes are embedded into the iso-surface of network output~\cite{park2019deepsdf,chen2019implicit,Occupancy_Networks,zheng2021dit,Genova2020Local,Bozic2021NeuralDeformationGraph,deng2019NASA}. Another line of work on implicit representation aimed at learning scene representations for novel view synthesis from posed 2D images. They represent static scenes using voxel grids of high-dimensional features~\cite{sitzmann2019deepvoxels}, continuous learnable function~\cite{sitzmann2019srns} or neural radiance fields (NeRF)~\cite{mildenhall2020nerf}. NeRF, in particular, shows strong capability of modeling view-dependent effects and thus attracts much attention~\cite{liu2020nsvf,MartinBrualla20arxiv_nerfw,Zhang20arxiv_nerf++,reiser2021kilonerf,yu2021plenoctrees,Garbin21arxiv_FastNeRF,Lombardi2021MVP}. It is later extended for dynamic scenes through deformation learning~\cite{Park20arxiv_nerfies,Pumarola20arxiv_D_NeRF,Gafni20arxiv_DNRF,Tretschk20arxiv_NR-NeRF,Li20arxiv_nsff,Xian20arxiv_stnif,Gao-freeviewvideo,li2021neural3dvideo,shao2022doublefield}. Human motions are usually much more challenging to learn using neural networks, and several works~\cite{peng2021neuralbody,noguchi2021narf,kwon2021neural} incorporated prior from a statistical body template to tackle this difficulty. Note that most of these works can only playback the dynamic sequence that the networks are trained on, while our work aims at animation, which is a much harder task because the method has to generalize to new poses.

\textbf{Animatable Human Avatars.} 
In the last decade, many efforts have been made for achieving expressive and animatable 3D models for human avatars. To facilitate geometric learning, several statistical parametric templates are developed for face~\cite{FLAME:SiggraphAsia2017}, hands~\cite{Romero2017MANO,Moon_2020_ECCV_DeepHandMesh} and minimally clothed body~\cite{loper2015smpl,STAR:2020,SMPL-X:2019,TotalCapture2018}. 
To acquire animatable characters wearing casual clothes, traditional pipelines mostly reconstruct a subject-specific mesh template in advance, and then generate its motions using physics simulation~\cite{Stoll2010Videobased,DRAPE2012}, deformation space modeling~\cite{TotalCapture2018}, or deep learning~\cite{habermann2021realtimeDDC,timur2021driving_signal,Xiang2021ModelingClothing}. The reliance on pre-scanning efforts can be eliminated via deforming a general body template, and several works proposed to directly learn this deformation from geometric data~\cite{CAPE:CVPR:20,pons2017clothcap,Ma:SCALE:2021,Ma:POP:2021} or RGB videos~\cite{alldieck2018videoavatar,alldieck2018videoavatar_detailed,alldieck2019tex2shape,alldieck19octopus}. The texture map and the rasterization step in those methods are later replaced with neural texture maps and image decoders in order to achieve photo-realistic rendering~\cite{Liu2018Neural,liu2020NeuralHumanRendering,Shysheya2019TNR,raj2020anr}. Recently, neural scene representations and rendering techniques are adopted for higher-fidelity results \cite{peng2021animatable_nerf,peng2021neuralbody,neural_actors}. However, state-of-the-art methods only demonstrate results of tightly-fitting garments, while our method is more general in terms of clothes topology and deformation.

%% file: 3_overview.tex
\section{Representation}
\label{sec:overview}

Our goal is to learn an animatable virtual characters directly from RGB videos and to support loose clothes like skirts and dresses without pre-scanning a template. To this end, we propose a new representation that has a strong capability of modeling the shape, appearance and dynamic deformations of clothed humans. At its core is a set of \emph{structured local radiance fields}, each of which models the dynamic appearance inside a local space while moving according to the body poses as well as the cloth deformations. 
To be more specific, we first pre-define $N$ nodes on the SMPL model via farthest point sampling. Their coordinates on the canonical SMPL surface are denoted with $\{\bm{\bar{n}}_i\}_{i=1}^N$. Since the nodes are sampled from the SMPL model, each of them has an associated skinning weight vector $\bm{\omega}_i\in\mathbb{R}^J$, where $J$ is the number of body joints. 
Given a pose vector $\bm{\theta}^{(t)}$ at time stamp $t$, we can transform node $i$ to the posed space using linear blending skinning (LBS): 
\begin{equation}
\label{eqn:reprez:lbs}
    \bm{\mathrm{T}}_i^{(t)} = \sum \omega_{i,j} \bm{\mathrm{M}}_j(\bm{\theta}^{(t)}), 
\end{equation}
\begin{equation}
\label{eqn:reprez:skinning}
    \bm{n}_i^{(t)} = \bm{\mathrm{T}}_i^{(t)} \bm{\bar{n}}_i, 
\end{equation}
where $\bm{\mathrm{M}}_j(\bm{\theta}^{(t)}) \in SE(3)$ is the rigid transformation of the $j$-th body joints and $\omega_{i,j}$ is the $j$-th entry of $\bm{\omega}_i$. 

In Eqn. (\ref{eqn:reprez:skinning}), the nodes strictly follow the motion of the body surface. In order to handle the non-rigid deformations of clothes, we allow the nodes to shift independently. Mathematically, we assign a time-varying residual translation $\Delta\bm{n}_i^{(t)}$ to node $i$  in the canonical space, and modify Eqn. (\ref{eqn:reprez:skinning}) into:
\begin{equation}
\label{eqn:reprez:skinning_2}
    \bm{n}_i^{(t)} = \bm{\mathrm{T}}_i^{(t)} \left(\bm{\bar{n}}_i + \Delta\bm{n}_i^{(t)}\right).
\end{equation}

Finally, we construct a local radiance field over the influence of each node, with a function $\mathcal{F}_i$ represented by a tiny MLP. This MLP takes as input a coordinate in the local space of node $i$ and outputs a high-dimensional feature vector. To model the fine-grain dynamic details that cannot be represented by node translations, we condition the local radiance field on a dynamic detail embedding $\bm{e}_i^{(t)}$. Formally, given any point $\bm{p}\in\mathbb{R}^3$ in the posed space at frame $t$, we first calculate its coordinate in the local space of node $i$ as:
\begin{equation}
\label{eqn:reprez:local_coord}
    \bm{p}_i = \left(\bm{\mathrm{T}}_i^{(t)}\right)^{-1} \bm{p} - \left(\bm{\bar{n}}_i + \Delta\bm{n}_i^{(t)}\right). 
\end{equation}
After that, we feed it into the local radiance network $\mathcal{F}_i$ and blend the feature vectors produced by all local MLPs:
\begin{equation}
\label{eqn:reprez:feat_fusion}
    \bm{f} = \frac{\sum w_i \mathcal{F}_i(\bm{p}_i; \bm{e}_i^{(t)})}{\sum w_i},  
\end{equation}
where $w_i$ is the blending weight defined as
\begin{equation}
\label{eqn:reprez:blending_weight}
    w_i = \max \{\exp (-\Vert\bm{p} - \bm{n}_i^{(t)}\Vert_2^2 / 2\sigma^2) - \epsilon, 0\}, 
\end{equation}
and $\epsilon$ is a hyperparameter controlling the influence radius of the nodes.
This blended feature $\bm{f}$ is fed into two additional MLPs, $\mathcal{G}(\cdot)$ and $\mathcal{H}(\cdot)$, to compute the color \& density of $\bm{p}$:
\begin{equation}
\label{eqn:reprez:color_density}
    \mathrm{Color}(\bm{p}) = \mathcal{G}(\bm{f}, \bm{v}), \quad \mathrm{Density}(\bm{p}) = \mathcal{H}(\bm{f}), 
\end{equation}
where $\bm{v}\in\mathbb{R}^3$ is the viewing direction~\cite{mildenhall2020nerf}. 

Overall, the dynamic appearance of a clothed character is parameterized in a coarse-to-fine fashion with three sets of variables: body poses $\{\bm{\theta}^{(t)}\}$, node residual translations $\{\Delta\bm{n}_i^{(t)}\}$ and dynamic detail embeddings $\{\bm{e}_i^{(t)}\}$. With the radiance field determined by these variables and the networks (\textit{i.e.}, $\mathcal{F}_1, \mathcal{F}_2, ..., \mathcal{F}_N$, $\mathcal{G}$ and $\mathcal{H}$), we can shoot rays and render images via volume rendering as in \cite{mildenhall2020nerf}. An illustration of our representation is presented in Fig.~\ref{fig:reprez}.

\textbf{Discussion. }
Compared to state-of-the-art methods, our representation has two advantages:
\begin{itemize}[leftmargin=*]
\setlength{\itemsep}{0pt}
\setlength{\parsep}{0pt}
\setlength{\parskip}{0pt}
\vspace{-0.2cm}
    \item Our method has expressive representation power in terms of both the motion and the topology. Although the nodes in our representation are sampled from the SMPL model, our method is not restricted by it. Instead, our method allows more degrees of freedom for motion and geometry modeling, enabling avatar creation for different cloth topologies, which is a significant departure from the existing works \cite{peng2021animatable_nerf,neural_actors,Shysheya2019TNR,raj2020anr}. 
    \item Our method does not explicitly define a global canonical field and consequently avoids the need for ``backward skinning" during training. Backward skinning is used to transform the points in the posed space to a global canonical space, and has been the basis of previous methods  \cite{peng2021animatable_nerf,neural_actors,Saito:SCANimate:2021}. Even so, we argue that this operation is ambiguous, especially for the points around contacting body parts. In contrast, our approach computes the radiance of any point in the local space, thus resolving the ambiguity issue. 
\vspace{-0.2cm}
\end{itemize}

\begin{figure}
    \centering
    \includegraphics[width=1.0\linewidth]{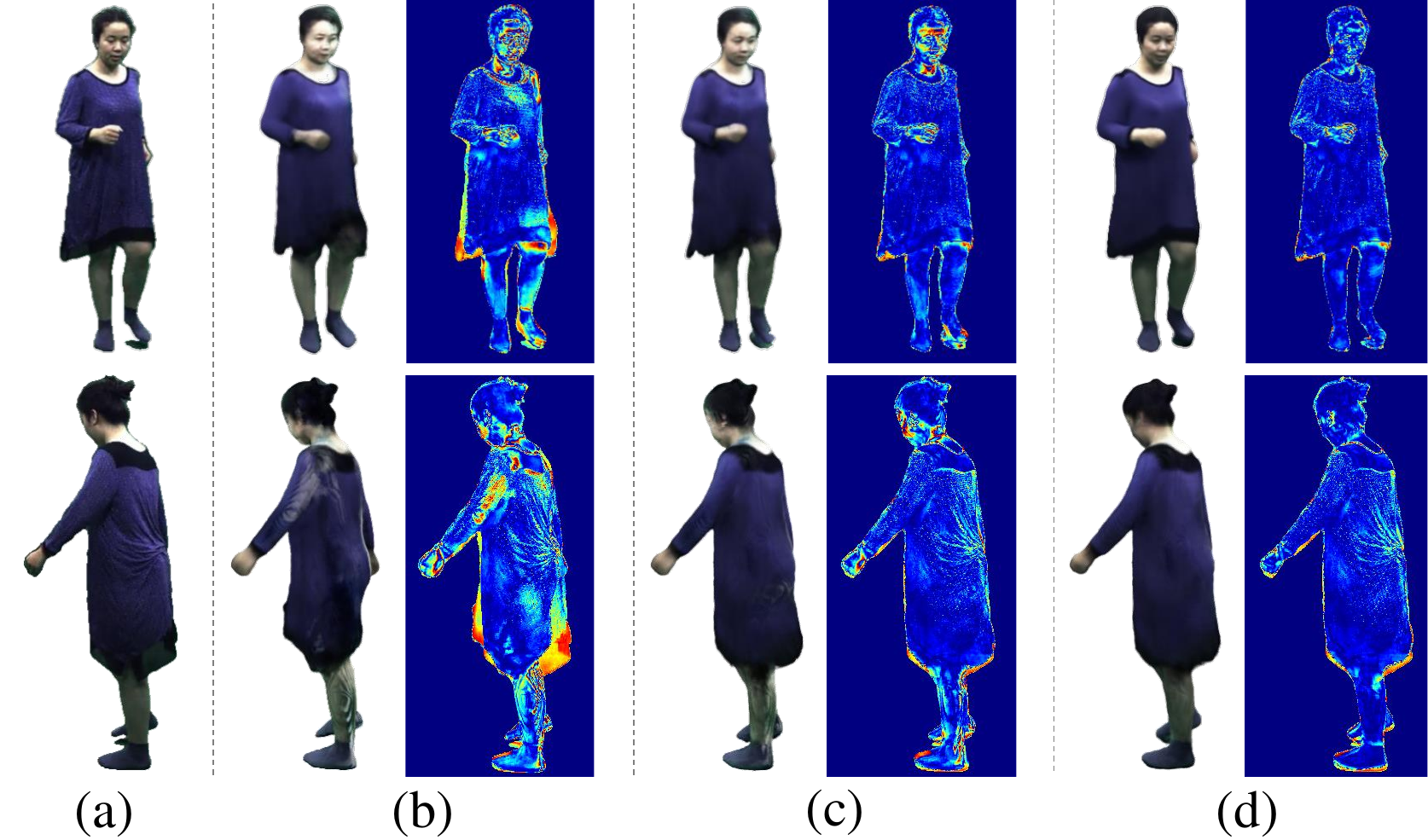}
    \caption{\textbf{Visualization of the effect of node-related variables.} (a) Ground-truth reference. (b) Rendering results without node residual translation and dynamic detail embeddings. (c) Results without dynamic detail embeddings. (d) Results with full set of variables. See Sec.~\ref{sec:experiments:evaluation} for details.}
    \label{fig:eval_node_effect}
\end{figure}

%% file: 4_method.tex

\section{Method}
\label{sec:method}
Having elaborated on the proposed representation, we turn to network learning in this section. 
Specifically, we need to determine the aforementioned variables alongside with the weights of the radiance networks for a training image sequence $\bm{I}_t, t=1, 2,..., T$. The images can be captured from a multi-view system or a monocular one. In order to synthesize images for new poses, we also have to compute the node residual translations and the dynamic detail embeddings corresponding to those poses. 
We assume access to the body poses of the training images (\textit{i.e.}, $\bm{\theta}^{(t)}, t=1, 2, ..., T$), which can be estimated using marker-less MoCap tools such as \cite{easymocap,lightcap2021}. The node residual translations and the detail embeddings are referred to as ``node-related variables" in the following context.



\subsection{Network Architecture}
\label{sec:method:arch}
To obtain the node-related variables for the training frames and ensure generalization during animation, we design a simple conditional variational auto-encoders (cVAE)~\cite{Sohn2015cvae} as an auxiliary network for each node. 
Each auxiliary network consists of an encoder and a decoder, both implemented with tiny MLPs. 
Following the practice of SCANimate~\cite{Saito:SCANimate:2021}, the condition variable of this cVAE is the pose vector multiplied by the skinning weight and an attention map:
\begin{equation}
    \bm{\theta}^{(t)}_i = (\bm{\mathrm{W}}\cdot \bm{\omega}_i) \circ \bm{\theta}^{(t)}, 
\end{equation}
where $\bm{\mathrm{W}}$ is the weight map that converts the skinning weights into pose attention weights as in \cite{Saito:SCANimate:2021} and $\circ$ denotes element-wise product. 
During training, the encoder takes the time stamp $t$ as input and $\bm{\theta}^{(t)}_i$ as condition, and produces parameters of a Gaussian distribution, from which a latent code $\bm{z}_i^{(t)}$ is sampled:
\begin{equation}
\label{eqn:latent_sample}
    \bm{\mu}_i^{(t)}, \bm{\sigma}_i^{(t)} \gets \mathcal{E}(t, \bm{\theta}_i^{(t)}), \ \bm{z}_i^{(t)} \sim \mathcal{N}(\bm{\mu}_i^{(t)}, \bm{\sigma}_i^{(t)}), 
\end{equation}
Conditioned on the body pose, the latent code is then decoded into the node residual translation and the dynamic detail embedding:
\begin{equation}
    \Delta\bm{n}_i^{(t)},  \bm{e}_i^{(t)} \gets \mathcal{D}(\bm{z}_i^{(t)}, \bm{\theta}_i^{(t)}), 
\end{equation}
which are later used in Eqn.~(\ref{eqn:reprez:local_coord}) and  Eqn.~(\ref{eqn:reprez:feat_fusion}), respectively. 

\begin{figure}
    \centering
    \includegraphics[width=1.0\linewidth]{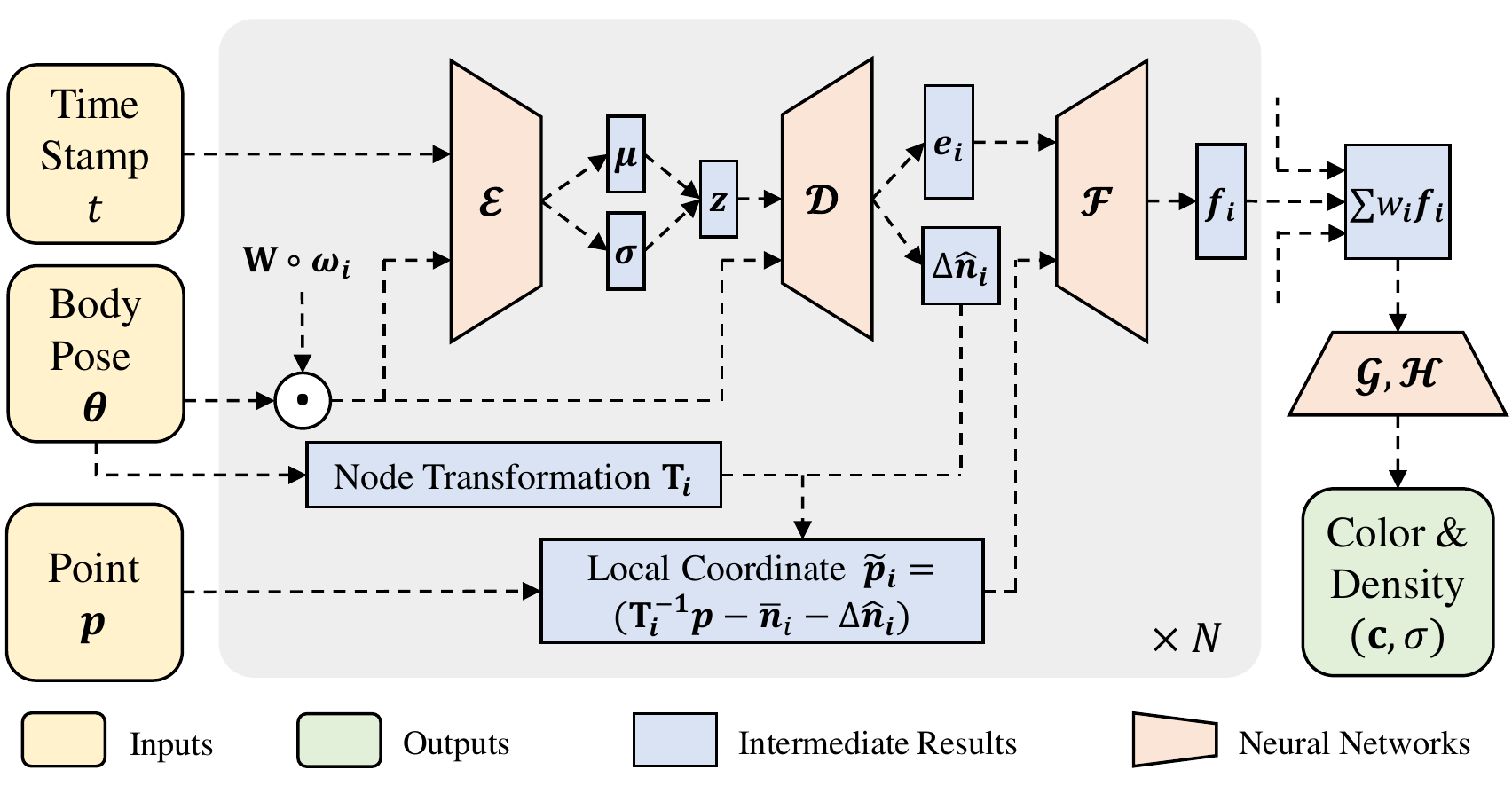}
    \caption{\textbf{Illustration of the data flow in our network.} The time stamp and body pose feature are first passed through the cVAEs, which produces the node residual translations and dynamic detail embeddings of the local radiance fields. For a point in the posed space, we calculate its local coordinate in each local field, and then query its feature. Finally, all features are blended and decoded into the color and density values.   }
    \label{fig:network}
\end{figure}

In this network, the time instant is used to distinguish similar poses at different time instants, thereby avoiding the one-to-many mapping issue. With the KL-divergence loss in cVAE, there is a preference to let the decoder to mainly rely on the pose condition for prediction, and the time input only provides information necessary for good reconstruction.  
In our implementation, we augment the time stamp and the coordinates with Fourier encoding before feeding them into MLPs~\cite{mildenhall2020nerf} . Fig.~\ref{fig:network} illustrates the data flow in our network during training. Once the training is done, we can render the model for either training frames or novel poses. To render the training sequence, we use the full network and set $\bm{z}_i^{(t)} = \bm{\mu}_i^{(t)}$ in Eqn.~(\ref{eqn:latent_sample}) to eliminate randomness. When unseen poses are given, the encoder half of the cVAE will be omitted and $\bm{z}_i^{(t)}$ will be set to zeros.

\subsection{Training Loss}
\label{sec:method:loss}
Our network can be trained in an end-to-end manner. 
The training loss is composed of four components, including a reconstruction loss, a node translation regularization, an embedding regularization, and a KL-divergence loss:
\begin{equation}
     \mathcal{L} = \lambda_{rec}\mathcal{L}_{rec} + \lambda_{trans}\mathcal{L}_{trans} + \lambda_{ebd}\mathcal{L}_{ebd} + \lambda_{KL}\mathcal{L}_{KL}.
\end{equation}
Below we discuss them in details. For ease of notation, we drop the superscript ${}^{(t)}$ of all variables in this subsection. 

\textbf{Reconstruction Loss} $\mathcal{L}_{rec}$ measures the mean squared error between the rendered and true pixel colors:
\begin{equation}
    \mathcal{L}_{rec} = \sum_{\bm{r}\in\mathcal{R}} \left\Vert \bm{C}\left(\bm{r} | \bm{\theta}, \{\Delta\bm{n}_i\}, \{\bm{e}_i\}\right) - \hat{\bm{C}}_{\bm{r}}  \right\Vert_2^2, 
\end{equation}
where $\mathcal{R}$ is the set of rays in each batch, $\hat{\bm{C}}_{\bm{r}}$ is the ground-truth pixel color, $\bm{C}(\cdot | \bm{\theta}, \{\Delta\bm{n}_i\}, \{\bm{e}_i\})$ is the volume rendering function with the representation defined in Sec.~\ref{sec:overview}. 

\textbf{Node Translation Regularization} $\mathcal{L}_{trans}$ simply constrains the position change of each nodes in order to stabilize training: 
\begin{equation}
    \mathcal{L}_{trans} = \sum_i \Vert \Delta\bm{n}_i \Vert_2^2. 
\end{equation}

\textbf{Embedding Regularization} $\mathcal{L}_{ebd}$ penalize large magnitudes of the dynamic detail embeddings: 
\begin{equation}
    \mathcal{L}_{ebd} = \sum_i \Vert \bm{e}_i \Vert_2^2. 
\end{equation}
A similar loss is also used in \cite{park2019deepsdf}; here we utilize it to encourage the embeddings to encode only the information that cannot be represented by node position. 

\textbf{KL-divergence Loss} $\mathcal{L}_{KL}$ is a standard VAE KL-divergence penalty~\cite{kingma2014vae}:
\begin{equation}
    \mathcal{L}_{KL} = \sum_i \mathrm{KL}\left( \mathcal{N}(\bm{\mu}_i, \bm{\sigma}_i) \ \Vert\  \mathcal{N}(\bm{0}, \bm{\mathrm{I}})\right).
\end{equation}


\noindent\textbf{Implementation Details}
The local radiance networks and cVAEs in our architecture are implemented with parallel tiny MLPs in the form of group 1D convolution.
To accelerate training and inference, we exploit the fact that, for any point in the posed space, only a small portion of nodes have influence on its color and density value. 
We use Adam optimizer to train our models. Training the whole models takes about 25 hours on one NVIDIA 3090 GPU with 500k iterations, while rendering an color image with resolution of $512\times 512$ typically takes 5 seconds on one NVIDIA 3080TI GPU. Please refer to the \textit{Supp.Mat.} for more details.

%% file: 5_results.tex
\begin{figure*}
    \centering
    \includegraphics[width=0.98\linewidth]{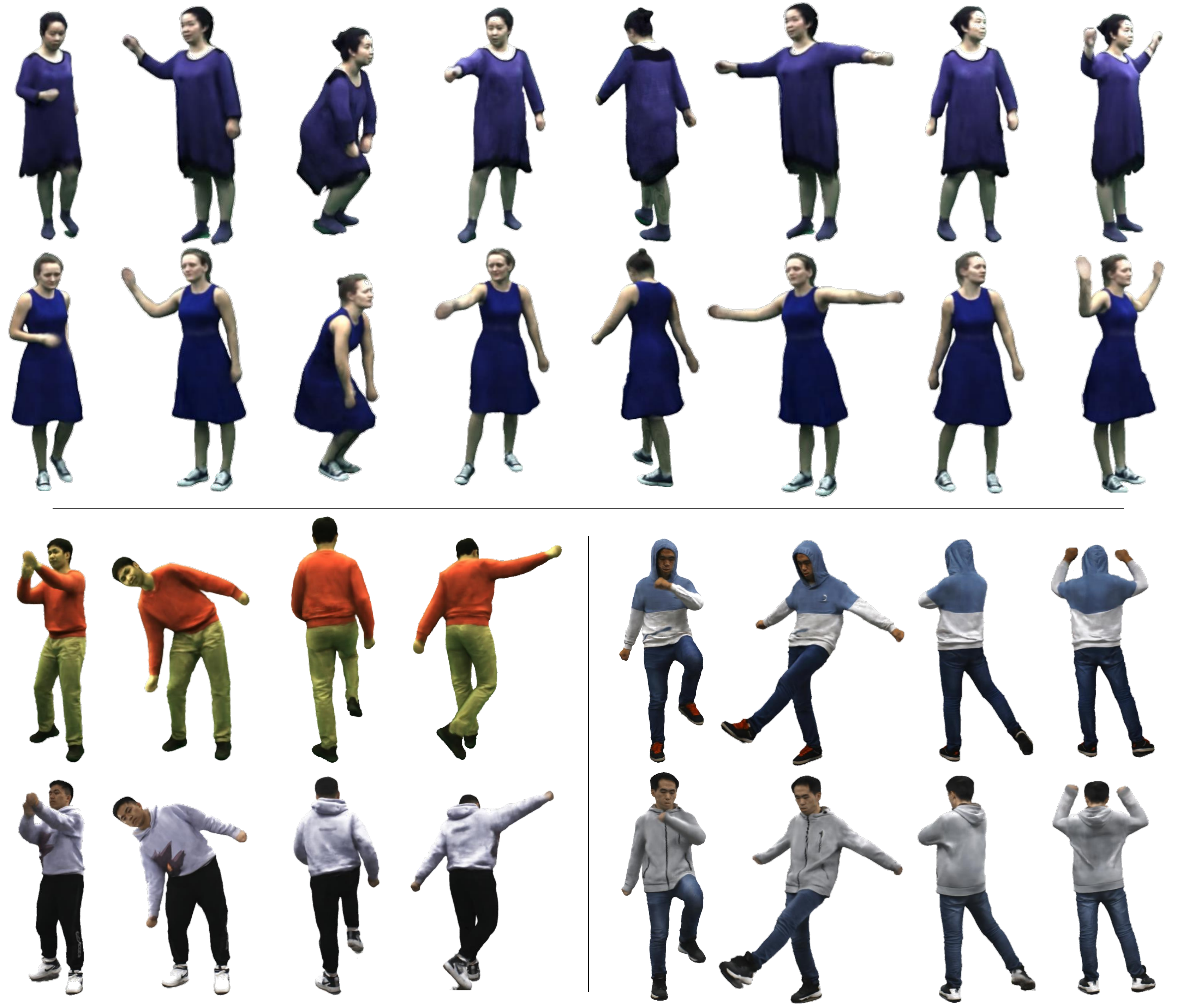}
    \caption{\textbf{Example results of our method.} We train our network on various datasets and show the novel pose synthesis results. }
    \label{fig:results}
\end{figure*}

\begin{figure}
    \centering
    \includegraphics[width=1.0\linewidth]{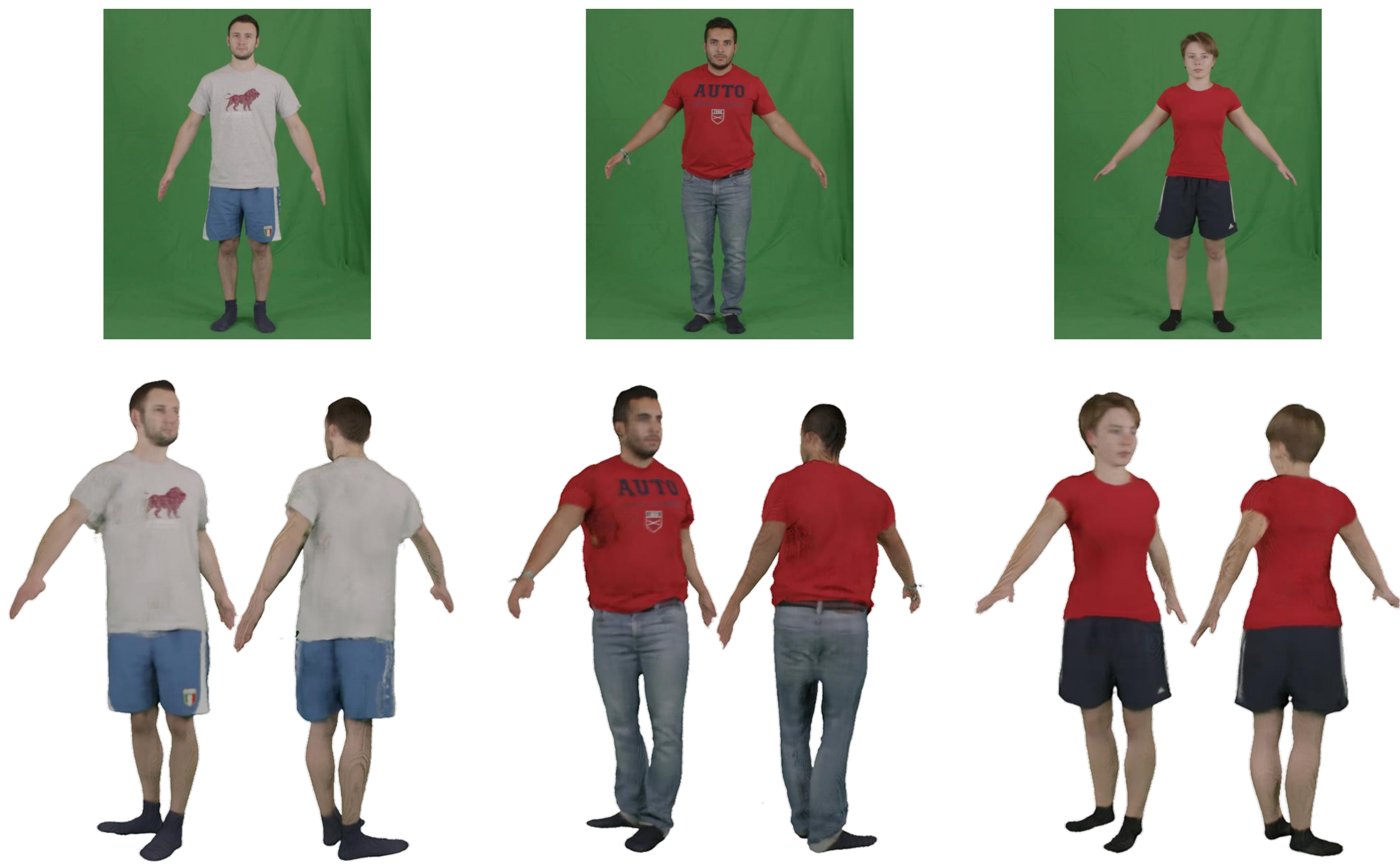}
    \caption{\textbf{Our results on PeopleSnapshot dataset.} Given a monocular video recording a person rotating in an A-pose (top), our method is able to create a human avatar that supports novel pose generation and free view synthesis (bottom). }
    \label{fig:singleview}
\end{figure}

\section{Experiments}
\label{sec:experiments}
\noindent\textbf{Dataset and Metrics. } 
For evaluation and comparison with baseline methods, we mainly use the following dataset: 
(1) Two dress sequences from \cite{habermann2021realtimeDDC}, which are captured using 100 cameras but we manually select 20 views among them for computational efficiency; (2) One sweater sequences from \cite{habermann2020deepcap} captured with 10 cameras; (3) Two sequences from ZJU-MoCap~\cite{peng2021neuralbody} captured with 23 cameras; and (4) three multi-view sequences collected by ourselves with 24 cameras\footnote{Data collection and disclosure have been consented by the volunteers. }. For quantitative evaluation, we use two standard metrics: peak signal-tonoise ratio (PSNR) and structural similarity index (SSIM).
More details about data collection and preprocessing can be found in the \textit{Supp.Mat.}.

\subsection{Results}
We train our model for each individual subjects, and present some example animation results in Fig.~\ref{fig:teaser} and Fig.~\ref{fig:results}. 
The results cover various body poses and different cloth styles. 
As shown in these figures, our method not only gracefully tackles different cloth types, but also generates realistic dynamic wrinkles. 
Please see our supplemental video for more visualization. 


Although we mainly use multi-view videos for evaluation, our method is also able to learn an avatar from single-view input. Fig.~\ref{fig:singleview} demonstrates the results of our method on  the PeopleSnapshot dataset~\cite{alldieck2018videoavatar}, which captures performers rotating 360 degrees in an A-pose with a monocular camera. As shown in the figure, our method can also work well with such extremely simple input, further proving its generalization capability.

\begin{figure}
    \centering
    \includegraphics[width=1.0\linewidth]{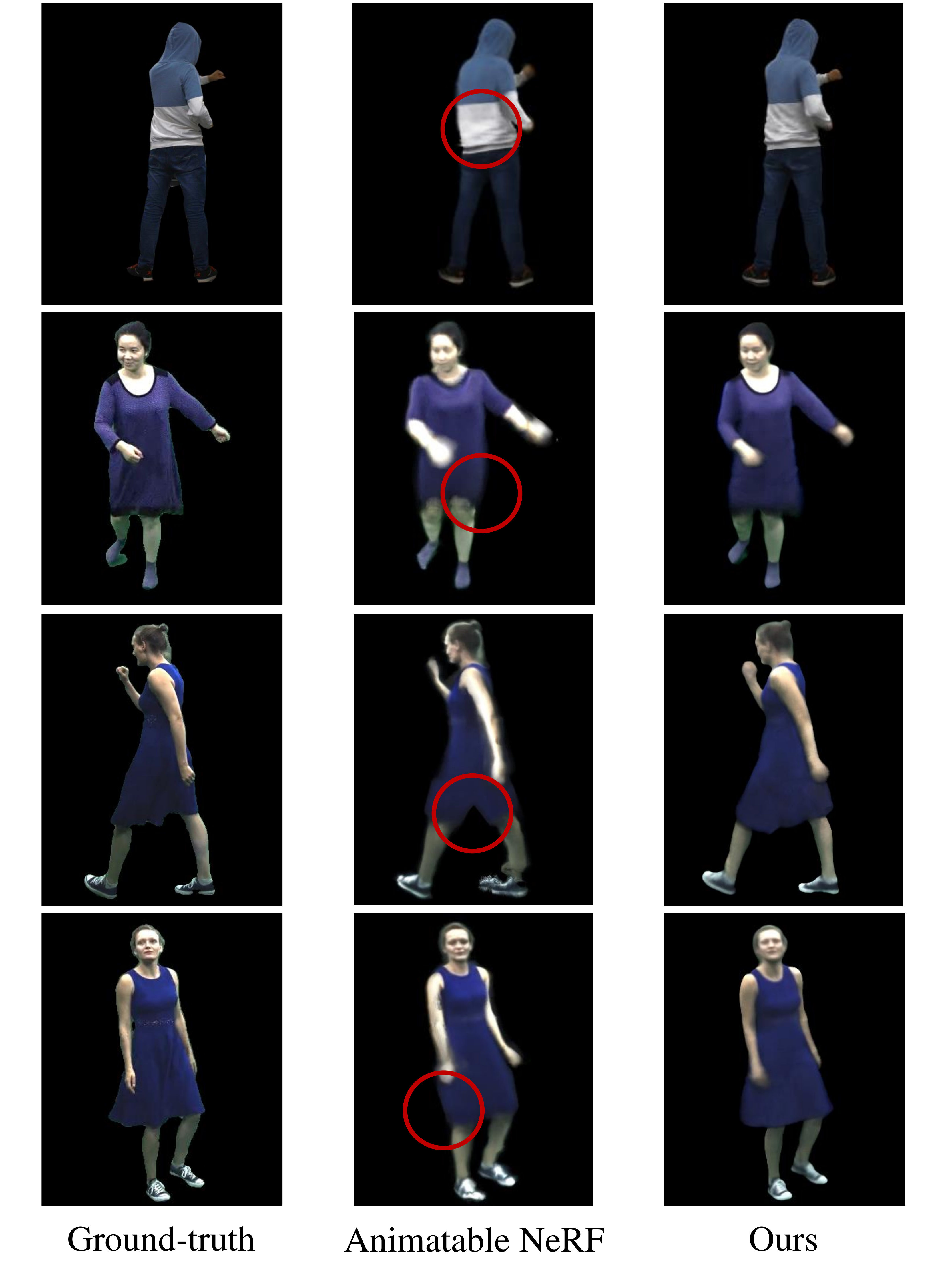}
    \caption{\textbf{Comparison against Animatable Nerf}~\cite{peng2021animatable_nerf} on novel pose synthesis. }
    \label{fig:anerf_comparison}
\end{figure}

\subsection{Comparison}
We mainly compare our method with Animatable NeRF~\cite{peng2021animatable_nerf} and Neural Body~\cite{peng2021neuralbody}. 
We omit other related methods since they have been compared in \cite{peng2021animatable_nerf}. 

We first compare with Animatable NeRF~\cite{peng2021animatable_nerf} on the dataset of \cite{habermann2021realtimeDDC} and our own data. 
We split each video into training frames and testing ones, train the networks using the training frames from all views, and test the animation quality using the testing frames. 
Qualitative results are presented in  Fig.~\ref{fig:anerf_comparison}. 
Compared to \cite{peng2021animatable_nerf}, our method can produce more appearance details, and generate the non-rigid motions of dress hems. 
The numeric results in Tab.~\ref{tab:anerf_comparison} also prove that our method can achieve higher-quality results than \cite{peng2021animatable_nerf}.

\begin{figure}
    \centering
    \includegraphics[width=1.0\linewidth]{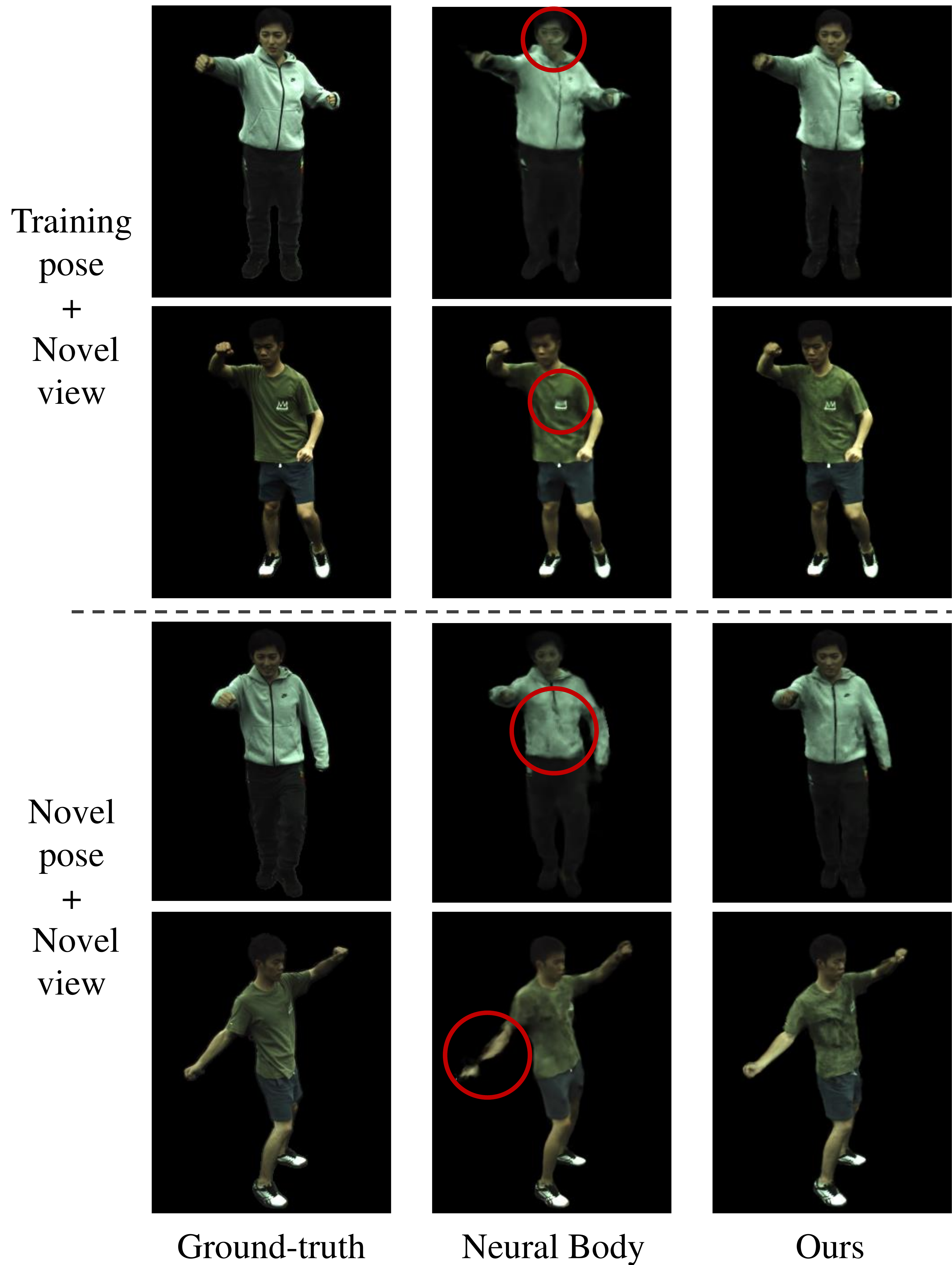}
    \caption{\textbf{Comparison against Neural Body}~\cite{peng2021neuralbody} in terms of both novel view synthesis and pose generation. Zoom in for better view.}
    \label{fig:neuralbody_comparison}
\end{figure}

To conduct a fair comparison with Neural Body~\cite{peng2021neuralbody}, we use their dataset and follow the same protocal in their paper. In this comparison, we train our network using only 300 image frames from four views, as done in \cite{peng2021neuralbody}. We evaluate the quality of novel view synthesis for training frames and unseen body poses. 
The results in Tab.~\ref{tab:neuralbody_comparison} shows that our model achieves higher accuracy than \cite{peng2021neuralbody} in both metrics. In fact, our method performs better not only in learning appearance details like the logo, but also in generalizing to unseen poses, as shown in Fig.~\ref{fig:neuralbody_comparison}. We also report the numeric results of Animatable NeRF~\cite{peng2021animatable_nerf} in Tab.~\ref{tab:neuralbody_comparison} for completeness.

\begin{table}[t]
\centering
    \caption{Quantitative comparison with Animatable NeRF~\cite{peng2021animatable_nerf} in terms of novel pose synthesis.  }
    \scriptsize
    \begin{threeparttable}
    \begin{tabular}{lcccc}
        \toprule
         & \multicolumn{2}{c}{PSNR ($\uparrow$)} & \multicolumn{2}{c}{SSIM ($\uparrow$)}                \\
        \cmidrule(r){2-3} \cmidrule(r){4-5} 
        Case $\backslash$ Method    & \cite{peng2021animatable_nerf} & Ours & \cite{peng2021animatable_nerf} & Ours \\
        \midrule 
        Hoody & 22.43 & \textbf{24.94} & 0.893 & \textbf{0.928} \\
        Jacket & 24.30 & \textbf{25.24} & 0.909 & \textbf{0.927} \\
        Dress1  & 19.52 & \textbf{23.43} & 0.848 & \textbf{0.891} \\
        Dress2 & 20.49 & \textbf{22.19} & 0.877 & \textbf{0.900} \\
        \bottomrule
    \end{tabular}
    \end{threeparttable}
    \vspace{2pt}
    \label{tab:anerf_comparison}
\end{table}

\begin{table}[t]
  \centering
  \scriptsize
  \caption{Quantitative comparison with Neural Body~\cite{peng2021neuralbody} and Animatable NeRF~\cite{peng2021animatable_nerf} on ZJU-MoCap dataset.}
    \begin{tabular}{ccrrrrrr}
    \toprule
          &       & \multicolumn{3}{c}{PSNR ($\uparrow$)} & \multicolumn{3}{c}{SSIM ($\uparrow$)} \\
    \cmidrule(r){3-5} \cmidrule(r){6-8} 
    \multicolumn{1}{c}{ID} & Pose Type & \multicolumn{1}{c}{\cite{peng2021neuralbody}} & \multicolumn{1}{c}{\cite{peng2021animatable_nerf}} & \multicolumn{1}{c}{Ours} & \multicolumn{1}{c}{\cite{peng2021neuralbody}} & \multicolumn{1}{c}{\cite{peng2021animatable_nerf}} & \multicolumn{1}{c}{Ours} \\
    \midrule 
    \multirow{2}{*}{387} & Seen  & 25.79 & 24.38     & \textbf{28.32}     & 0.928 & 0.903     & \textbf{0.953} \\
          & Unseen & 21.60 & 21.29     & \textbf{23.61}     & 0.870  & 0.860     & \textbf{0.905} \\
    \midrule 
    \multirow{2}{*}{392} & Seen  & 29.44 & 27.43 & \textbf{30.79} & 0.946 & 0.919 & \textbf{0.958} \\
          & Unseen & 25.76 & 24.59 & \textbf{26.74} & 0.909 & 0.889 & \textbf{0.927} \\
    \bottomrule
    \end{tabular}%
  \label{tab:neuralbody_comparison}%
\end{table}%

\subsection{Ablation Study}
\label{sec:experiments:evaluation}
In this subsection, we conduct three qualitative ablation experiments on the main components of our method design. We present the quantitative results as well as some additional experiments in the \textit{Supp.Mat.}. 

\noindent\textbf{Node-related variables.} 
To understand the effect of the node-related variables in our method, we take the trained model for a dress sequence and conduct experiment on it. Specifically, we render the images of training poses under three circumstances, \textit{i.e.}, 1) without node residual translations or dynamic detail embeddings, 2) with node residual translations but without dynamic detail embeddings, and 3) with both node translations and detail embeddings. 
The results are shown in Fig.~\ref{fig:eval_node_effect}. 
As the figure shows, when the node residual translations and the dynamic detail embeddings are both disabled, the model only recovers the articulated motions and fails to render the correct shape of the moving character. With solely the node residual translation enabled, the non-rigid deformation of the dress hem can be recovered, but the shading on the facial area is not consistent with the image evidence. Only with both the node residual translation and the dynamic detail embeddings enabled can all appearance details be faithfully reconstructed.


\begin{figure}
    \centering
    \includegraphics[width=1.0\linewidth]{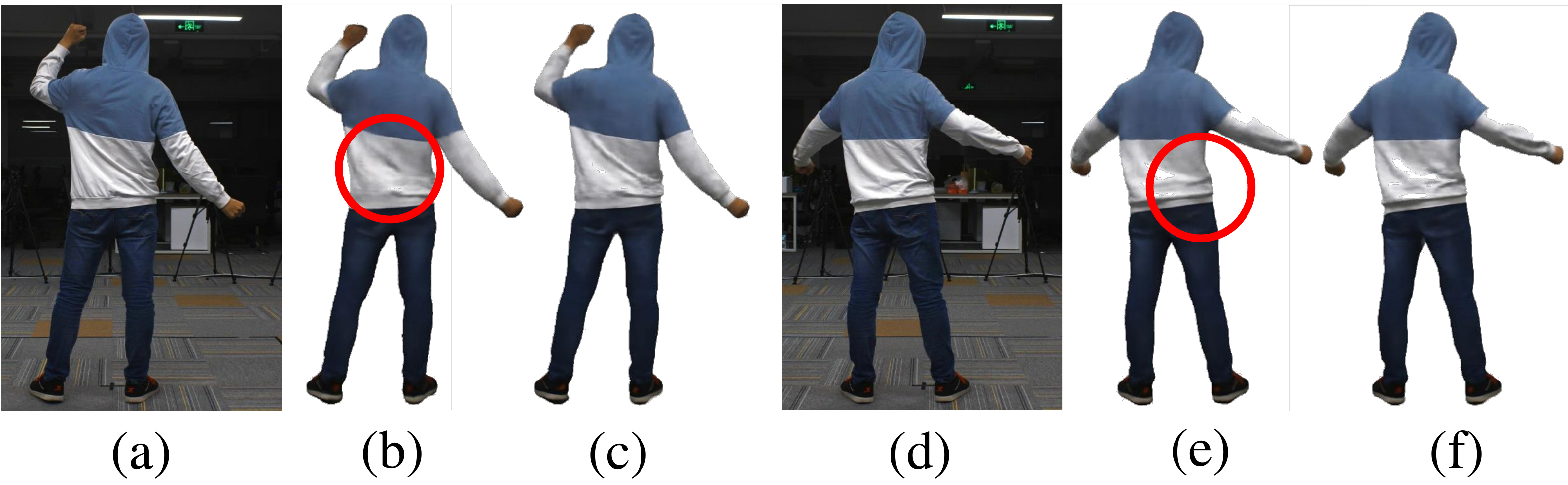}
    \caption{\textbf{Evaluation of our cVAE design.} We replace the cVAE with a determinstic regression network, and compare the reconstruction results of training frames. (a,d) Ground-truth. (b,e) Results by the deterministic baseline. (c,f) Our results.  }
    \label{fig:eval_cvae}
\end{figure}

\noindent\textbf{cVAE. }
We evaluate our choice of cVAE-based architecture by replacing it with a deterministic network that directly regresses the node-related variables from body poses. This baseline network is trained under the same setting as our proposed model. 
We render the images for training frames in order to compare the performance of data fitting, and the results are  presented in Fig.~\ref{fig:eval_cvae}. 
Not surprisingly, naively learning a mapping from pose parameters to the node-related variables, without specifically account for the potential one-to-many mapping problem, will produce averaged appearance and fail to recover the dynamic garment wrinkles even for training images. In contrast, our method can fit to training data much better than the baseline method, consequently enabling realistic animation and rendering.

\begin{figure}
    \centering
    \includegraphics[width=1.0\linewidth]{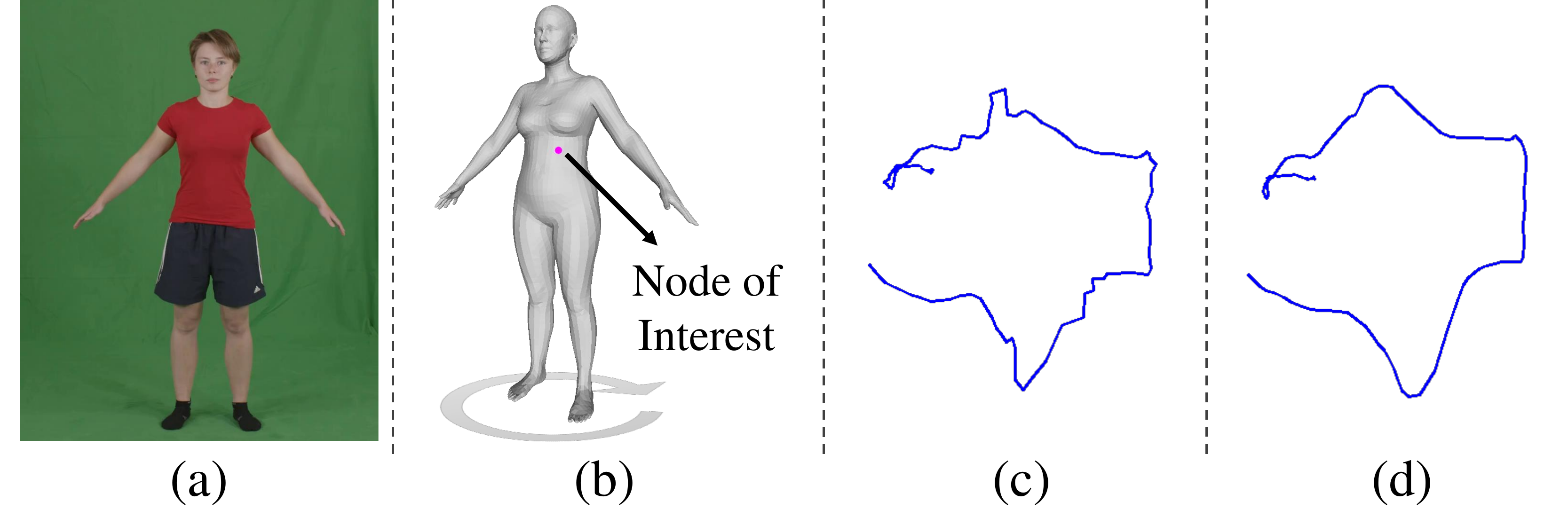}
    \caption{\textbf{Evaluation of the time instant input.} We replace the time stamp input with learnable per-frame latent codes and compare the trajectory of nodes. (a) Training video. (b) The node of which we visualize the trajectory. (c) Node trajectory using learnable latent codes. (d) Node trajectory using the proposed method. }
    \label{fig:eval_time}
\end{figure}

\noindent\textbf{Time stamp input. }
There exist other options that can be used as the cVAE input for resolving the one-to-many mapping problem. For instance, we can use learnable per-frame latent embeddings. The motivation behind our choice of time stamp is that, the low-frequency bias in MLPs can ensure temporal smoothness of the node-related variables, especially for node residual translations. In this way, we avoid the need for an additional loss of temporal smoothness. To validate this motivation, we conduct an ablation study where we replace the time stamp input with learnable latent embeddings. Then we compare the node trajectories as in Fig.~\ref{fig:eval_time}. As shown by the results, without explicitly constraining the temporal smoothness, the baseline method learns noisy node motions, while the trajectory of our method is much more smooth and physically plausible.

%% file: 6_conclusion.tex
\section{Discussion}
\label{sec:discussion}
\noindent\textbf{Conclusion.}
We introduced a novel method that uses structured local radiance fields for generation of controllable clothed human avatars. It has expressive representation power for both appearance and motion, as we leverage the advantages of neural scene representation while explicitly accounting for the motion hierarchy of clothes. Compared to existing methods, ours can handle more general cloth styles and generate realistic dynamic details.

\noindent\textbf{Limitation.}
The performance of our method depends on the pose variance in the training data, and our method may fail to generate plausible results when the animation poses starkly differ from the training poses; see Supp.Mat. for an example. 
In addition, the dynamic deformations and wrinkle changes of garments involve complex physics processes, which may be beyond the representation capability of our model. 
Finally, our method assumes accurate body pose estimation for the training images; that is why we mainly conduct experiments on multi-view dataset. For monocular videos, erroneous pose estimation caused by ambiguity may eventually lead to rendering artifacts.  

\noindent\textbf{Potential Social Impact.}
Our method enables automatic creation of a digital twin of any person. It can be combined with existing Deep Fake algorithms to generate fake videos through character animation and reenactment, which need to be addressed carefully before deploying the technology.

\noindent\textbf{Acknowledgement.}
This paper is sponsed by National Key R\&D Program of China (2021ZD0113503) and the NSFC No. 62125107 and No. 62171255.

%% file: 9_supp_context.tex

\section{Overview}
This supplementary document provides more discussions and experimental details. 
In Sec.~\ref{sec:more_discussion}, we discuss in detail the differences between our method and state-of-the-art approaches. 
Details about network architecture are presented in Sec.~\ref{sec:imp_details}. 
In Sec.~\ref{sec:exp_details} we present more details about how we collect the data and how we conduct the experiments. 
We conduct additional experiments in Sec.~\ref{sec:more_exp} to further evaluate our method design. 
Finally, we discuss the limitations and potential future work in Sec.~\ref{sec:limitation}. 
Please refer to the supplementary video for more visualizations. 

\section{More Discussion}
\label{sec:more_discussion}
Our method aims at creating a controllable 3D human character from RGB videos without pre-scanning a subject-specitic template. 
To better motivate our method and differentiate from existing approaches, we list the most related works below and discuss their limitations as well as our solution in this section. 

\textbf{Neural Body}~\cite{peng2021neuralbody} attaches learnable latent codes to the vertices of SMPL model, and employs sparse 3D convolutions to diffuse the latent codes into a radiance field in the 3D space. This scheme shows impressive performance on novel view synthesis for human performance. However, it struggles with new pose syntheses, as shown in \cite{peng2021animatable_nerf}. The main reason for this limitation is that 3D convolution is not equivalent to spatial changes of the structured latent code. In our method, we avoid the need for 3D convolutions and construct the radiance field by combining a set of localized ones, thus easily enable avatar animation by design.

\textbf{Animatable NeRF}~\cite{peng2021animatable_nerf} factorizes a deforming human body into a canonical radiance field and per-frame deformation fields that establish correspondences between the observations and the canonical space. 
The deformation field is generated through diffusing the input skeleton motion into the 3D space based on the learnable blending weights. 
Thanks to the explicit disentanglement of shape and motion, Animatable NeRF~\cite{peng2021animatable_nerf} is able to synthesize images for unseen poses. 
However, the motion representation is too simple to model the complex non-rigid deformations of clothes, which results into unrealistic, static texture and even severe artifacts when applying this method on loose clothes. In contrast, our method explicitly takes into account the non-rigid cloth deformation via coarse-to-fine decomposition, and demonstrates plausible animation results for human characters wearing dresses.

\textbf{Neural Actor}~\cite{neural_actors} shares a similar scheme with Animatable NeRF~\cite{peng2021animatable_nerf}: it also learns a neural radiance field in a canonical body pose, and use LBS to warp the canonical radiance field to represent the moving subject. Its main innovations are two fold: 1) Neural Actor learns pose-dependent non-rigid deformation that cannot be captured by standard skinning using a residual function, and 2) Neural Actor encodes appearance features on the 2D texture maps of the SMPL model to better capture dynamic details. Although this scheme shows impressive results in modeling the pose-dependent appearance details like the cloth wrinkles, it only works well for clothing that is topologically similar to the body. Besides, Neural Actor~\cite{neural_actors} requires multi-view input in order to obtain a complete texture map for network training. 
Note that we also use SMPL model in our approach; but we do not explicitly depend on the SMPL topology for shape and appearance representation. Therefore, our method is more general than Neural Actor~\cite{neural_actors} in terms of the cloth topology, and can work with partial input such as a monocular video.

\textbf{DDC}~\cite{habermann2021realtimeDDC} is another state-of-the-art method for building animatable avatars. It demonstrates impressive results for loose clothes and even achieves real-time rendering performance. However, DDC requires a pre-scanned template model of the actor; that is why we do not compare with it since person-specific templates are not available in our experiment setting. In contrast, our method can model the dynamic shape and appearance of general garments without any pre-scanning efforts.


Some methods like \textbf{TNR}~\cite{Shysheya2019TNR} and \textbf{ANR}~\cite{raj2020anr} learn animatable avatars in 2D domain. They typically define appearance features (RGB color values or high-dimensional features) on the UV map of a body template, and exploit a 2D convolutional network to obtain the final color image. These methods not only suffer from the same limitation as \cite{peng2021animatable_nerf,neural_actors}, but also fail to guarantee view consistency when rendering free-viewpoint images. Our method focuses more on creating a \textbf{3D} model, thereby significantly departing from this line of works.

\textbf{MVP}~\cite{Lombardi2021MVP} also proposes to use local volumetric representation for deformable surface rendering. However, our work is essentially different from MVP:
1) MVP requires an estimate of scene geometry to construct the volumetric primitives, while our method works without knowing scene geometry; 
2) MVP assumes accurate tracking of scene geometry over time, while our method is carefully designed to directly learn the motion hierarchy from data; 
3) MVP only handles head movements and facial expressions, while our method can deal with challenging body motions and cloth deformations;  
4) MVP mainly focuses on efficient rendering of training frames, while our method supports novel pose generation with explicit pose control.



\section{Implementation Details}
\label{sec:imp_details}
\subsection{Architecture Details}

\begin{figure}
    \centering
    \includegraphics[width=1.0\linewidth]{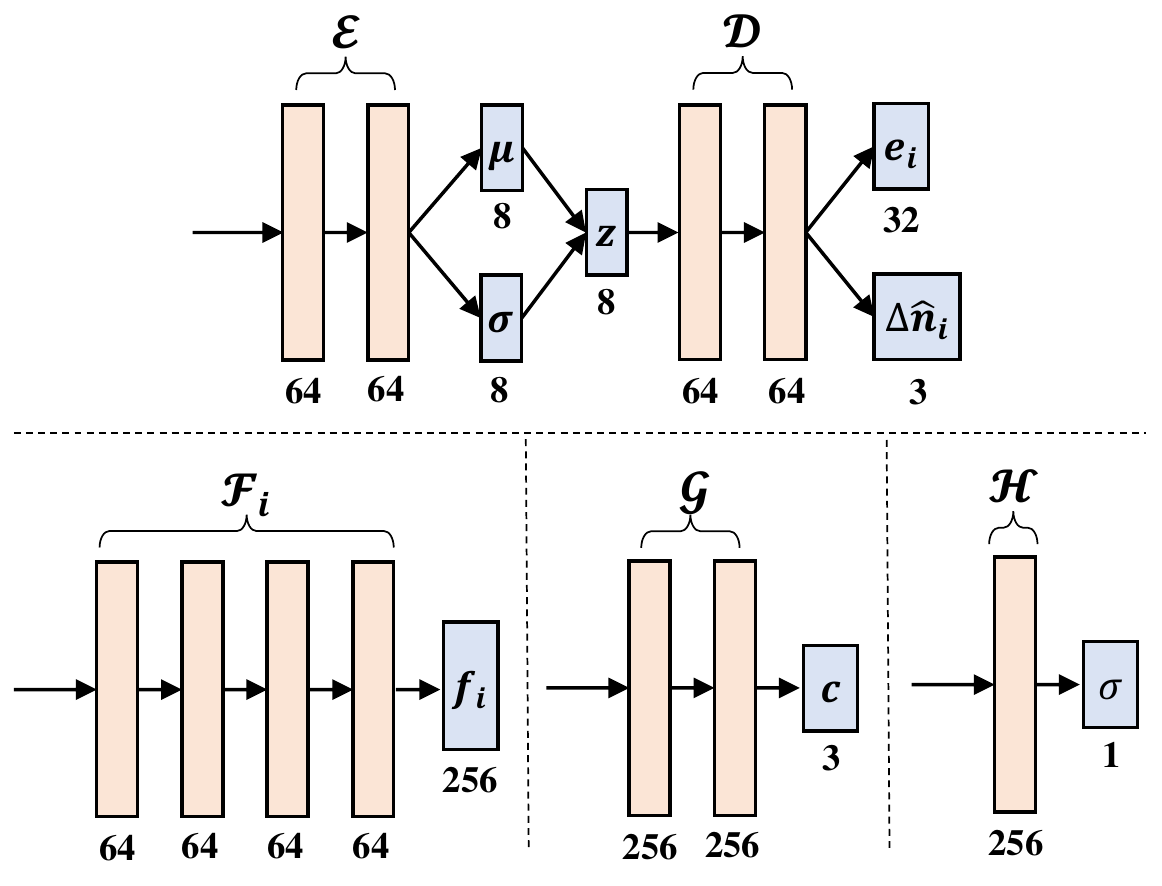}
    \caption{\textbf{Architecture of our network.} Each orange rectangle represents a fully-connected layer followed by ReLU activation, and the numbers of output channels are labeled underneath. }
    \label{fig:arch_details}
\end{figure}

We illustrate the network architecture in Fig.~\ref{fig:arch_details}. Note that before feeding the coordinates, view directions and time stamps into the MLP, we augment them using sinusoidal encoding, which is defined as: 
\begin{equation}
    \gamma (\bm{\mathrm{x}}) = \left(\bm{\mathrm{x}}, \sin(\bm{\mathrm{x}}), \cos(\bm{\mathrm{x}}), ..., \sin(2^{m-1}\bm{\mathrm{x}}), \cos(2^{m-1}\bm{\mathrm{x}}) \right). \nonumber
\end{equation}
The value of $m$ is 6 for coordinates, 4 for view directions and 12 for time stamps. We normalize the time stamp before sinusoidal encoding, \textit{e.g.}, the time stamp for $t$-th frame is normalized to $t/T$, where $T$ is the total number of frames. 

Note that the vanilla NeRF adopts a hierarchical sampling strategy and  simultaneously optimizes two networks (one ``coarse" and one ``fine"), while we only train one network with uniform sampling for fair comparison against baseline methods. 

\subsection{Network Acceleration}
Naively implementing our network will lead to heavy computational complexity, as one needs to query every local network for all point samples. 
To reduce network queries and accelerate program execution, we exploit the fact that for any point in the posed space, only a small portion of nodes have an influence on its color and density value. This is because the influence range of the nodes is truncated, as mathematically defined in Eqn. 6.
Based on this observation, we implement custom CUDA kernels for acceleration purpose. 
To be more specific, let $S$ denote the number of point samples and $N$ the number of nodes (which is also the number of local MLPs). In the naive implementation, the points are first transformed into the local coordinate systems of the nodes, which results into a tensor of size $N\times S \times 3$ being fed into the network. In our optimized implementation, we first calculate the number of necessary point queries for each local MLP (indexed by $i$), which is denoted as $S_i$. 
Then we construct an empty tensor of size $N\times S' \times 3$, where $S' = \max\{S_1, S_2, ..., S_N\}$. By investigating the values of blending weights, we pick the valid elements in the original tensor and rearrange them into the new one, which is finally fed into the network. 
With our optimized implementation, the memory consumption decreases about 85\%, and the running time decreases by a factor of 4.

\begin{table}
    \footnotesize
    \centering
    \caption{Hyperparameters for network training and evaluation. }
    \begin{tabular}{lc}
    \toprule
       Parameter Name & Value \\
    \midrule
       $N$ (Number of Nodes)                        & $128$ \\
       $\sigma$ (In Eqn. 6)                         & $0.05$ \\
       $\epsilon$ (In Eqn. 6)                       & $0.001$ \\
       $\lambda_{rec}$ (In Eqn. 11)                  & $1.0$ \\
       $\lambda_{trans}$ (In Eqn. 11)                & $0.02$ \\
       $\lambda_{ebd}$ (In Eqn. 11)                  & $0.1$ \\
       $\lambda_{KL}$ (In Eqn. 11)                   & $1\times 10^{-5}$ \\
       Dimension of $\bm{e}_i$ (In Eqn. 5)          & $32$  \\
       Dimension of $\bm{z}_i$ (In Eqn. 9)          & $8$  \\
       Number of Ray Samples Per Batch              & $2048$ \\
       Number of Point Samples Per Ray              & $64$ \\
       Batch Size                                   & $4$ \\
       Learning Rate                                & $5\times 10^{-4}$  \\
       Adam $\beta_1$                               & $0.9$ \\
       Adam $\beta_2$                               & $0.999$ \\
    \bottomrule
    \end{tabular}
    \label{tab:training_details}
\end{table}

\section{Experimental Details}
\label{sec:exp_details}
\subsection{Dataset}
In our experiments, we mainly use the following dataset: 
\begin{itemize}[leftmargin=*]
\setlength{\itemsep}{0pt}
\vspace{-0.2cm}
    \item Dataset from \cite{habermann2021realtimeDDC}. We use two dress sequences (``Ling" and ``FranziBlue") in this dataset. Each sequence contains about 20000 training frames captured using 100 cameras, but we manually select 20 views among them for computational efficiency. 
    \item Dataset from \cite{habermann2020deepcap}. We use one sweater sequence (``Lan") in this dataset, which is captured from 11 cameras and contains about 30000 training frames.   
    \item ZJU-MoCap dataset~\cite{peng2021neuralbody}. We mainly conduct experiments on two sequences (``CoreView387" and ``CoreView392"). Each sequence contains about 300 training frames captured from 23 view points, but we only use 4 view points among them for fair comparison against Neural Body~\cite{peng2021neuralbody}.  
    \item Multi-view dataset collected by ourselves. We built up a multi-view system that consists of 24 uniformly distributed cameras. Our system can capture synchronized videos at 30Hz with a resolution of 1024$\times$1024. We collect data for three subjects and the frame numbers of videos range from 2500 to 5000. 
\vspace{-0.2cm}
\end{itemize}

We use \cite{lightcap2021} to register SMPL(-X) model to the video frames, and use BackgroundMattingV2~\cite{BGMv2} for foreground segmentation. 

\subsection{Training Details}

We use PyTorch to implement our networks. The hyperparameters needed for network implementation and training are reported in Tab.~\ref{tab:training_details}. Note that during network training, the learning rate decays exponentially every 20k iterations. The number of iterations is set to 100k for People Snapshot dataset~\cite{alldieck2018videoavatar}, 300k for ZJU-MoCap dataset~\cite{peng2021neuralbody} and 500k for other multi-view sequences. 
For baseline methods, we use the author-provided code and run all the experiments using the default training settings.

\subsection{Metrics}
As described in the paper, we use two standard metrics, peak signal-to-noise ratio (PSNR) and structural similarity index (SSIM), for quantitative evaluation. To reduce the influence of background pixels, all the scores are calculated from the images cropped with a 2D bounding box which is estimated from the projection of SMPL model. More details are described in \href{https://github.com/zju3dv/neuralbody/blob/master/supplementary_material.md}{this link}.



\section{More Experiments}
\label{sec:more_exp}

\begin{table}
    \newcommand{\tabincell}[2]{\begin{tabular}{@{}#1@{}}#2\end{tabular}}
    \centering
    \scriptsize
    \caption{\textbf{Quantitative evaluation of the node-related variables.} We generate the images for training poses under different settings and report the averaged PSNR scores of all frames. }
    \label{tab:quant_node_effect}
    \scriptsize
    \begin{tabular}{lccc}
        \toprule
        Setting& w/o $\{\Delta\bm{n}_i^{(t)}\}$ or $\{\bm{e}_i^{(t)}\}$ & w/o $\{\bm{e}_i^{(t)}\}$ & Full  \\
        \midrule
        \tabincell{l}{Corresponding\\figure} & Fig.~3 (b) & Fig.~3 (c) & Fig.~3 (d)  \\
        \midrule
        PSNR & 17.52 & 20.47 & 21.48 \\ 
        \bottomrule
    \end{tabular}
\end{table}

\begin{table}
    \centering
    \scriptsize
    \caption{\textbf{Quantitative evaluation of our cVAE design.} We replace the cVAE with a determinstic regression network and report the reconstruction accuracy (PSNR) of training frames. }
    \label{tab:quant_cvae}
    \scriptsize
    \begin{tabular}{lcc}
        \toprule
        Setting& Deterministic & Ours  \\
        \midrule
        Corresponding figure & Fig.~9 (b,e) & Fig.~9 (c,f)  \\
        \midrule
        PSNR & 25.34 & 25.62 \\ 
        \bottomrule
    \end{tabular}
\end{table}

\noindent\textbf{Quantitative ablation of node-related variables.} We conduct a qualitative ablation study on the effects of the node-related variables in Fig.~3 in the main paper. In Tab.~\ref{tab:quant_node_effect} we report the corresponding quantitative results across all frames to further evaluate the impact of the node-related variables.

\noindent\textbf{Quantitative ablation of our cVAE design.} Similarly, we report the corresponding quantitative results across all frames in Tab.~\ref{tab:quant_cvae} to further evaluate our cVAE design. The numeric results further prove that our cVAE design is critical for better reconstructing the realistic details in the training frames, which is consistent with our conclusion in the main paper.

\begin{wrapfigure}{r}{0.35\linewidth}
\vspace{-2mm}
\hspace{-5mm}
\includegraphics[width=1.01\linewidth]{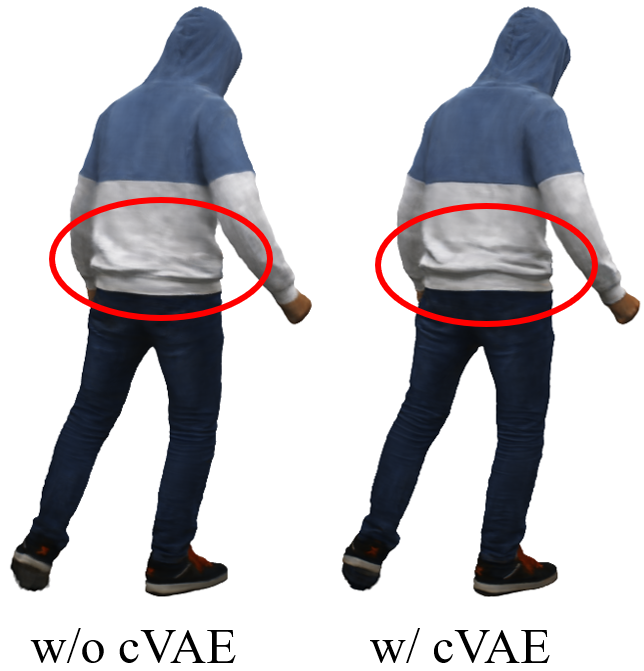}
\end{wrapfigure}
\noindent\textbf{cVAE ablation with novel poses.} 
In Fig.~9 we mainly conduct the cVAE ablation study on training frames. 
In the right inset figure we conduct an identical experiment using novel poses from a testing sequence. The results also show that our cVAE design is benefitial for learning sharper wrinkle details.

\begin{figure}
    \centering
    \includegraphics[width=1.0\linewidth]{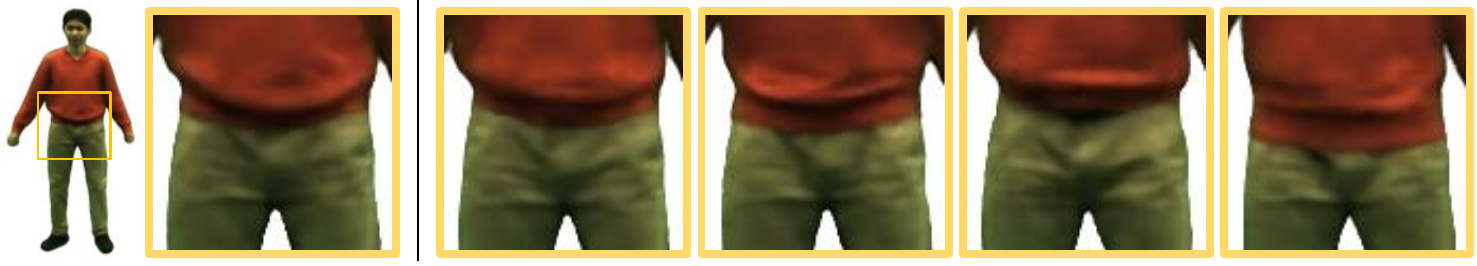}
    \caption{\textbf{Impact of the latent $\bm{z}_i$.} We show the testing results when the same pose are given but $\bm{z}_i$ is set to zero (leftmost) or assigned with random Gaussian noises (right). }
    \label{fig:latent_z}
\end{figure}

\noindent\textbf{Impact of the latent $\bm{z}_i$.} As we mentioned in Sec.~4.1, we set $\bm{z}_i$ to zeros when synthesizing images of novel poses. In fact, latent $\bm{z}_i$ does not have to be zeros and can be modified in accordance of applications. In Fig.~\ref{fig:latent_z}, we show that modifying the latent $\bm{z}_i$ will lead to different wrinkle patterns. This feature can be further explored to generate multiple plausible animation sequences, and we leave it as future work.


\section{Limitation and Future Work}
\label{sec:limitation}

\begin{wrapfigure}{r}{0.4\linewidth}
\vspace{-2mm}
\hspace{-5mm}
\includegraphics[width=1.01\linewidth]{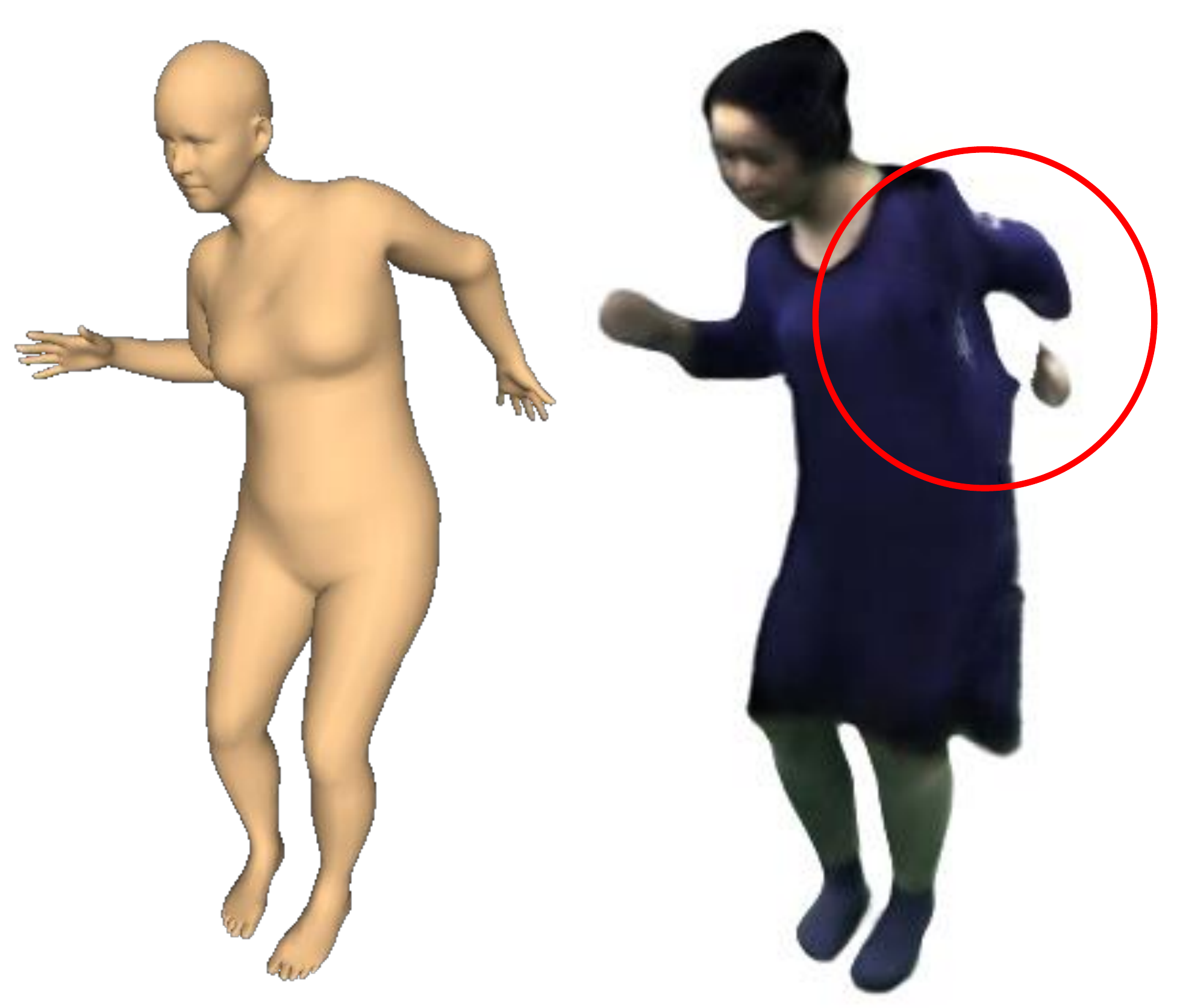}
\end{wrapfigure}
As we discuss in Sec.6 in the main paper, our method may fail to generate plausible results when the animation poses starkly differ from the training poses; see the inset figure on the right for an example. The main reason for this phenomenon is that neither subject-specific templates nor the SMPL surface is used to regularize shape learning in our method. 
Consequently, we cannot guarantee that our model is fully aware of the underlying geometry and its articulated surface deformation. 
Geometrical priors of clothed humans~\cite{Ma:POP:2021} can be employed to resolve this limitation and we leave it for future work. 

In addition, the dynamic deformations and wrinkle changes of garments involve complex physics processes, which may be beyond the representation capability of sparse nodes. Denser nodes could probably alleviate this limitation, but this will result in heavier computational burden. In fact, modeling the physics attributes of real-world garments is a long-standing, extremely difficult problem in computer graphics. We are currently seeking a better approach that can combine the merits of learning-based implicit representations and physics-based cloth simulation.